\documentclass[letterpaper]{article} 
\usepackage{aaai25}  
\usepackage{times}  
\usepackage{helvet}  
\usepackage{courier}  
\usepackage[hyphens]{url}  
\usepackage{graphicx} 
\usepackage{natbib}  
\usepackage{caption} 
\usepackage{algorithm}
\usepackage{algorithmic}
\usepackage{amsmath}
\usepackage{graphicx}
\usepackage{subfigure}
\usepackage{multirow}
\usepackage{booktabs}
\usepackage{xcolor}

\usepackage{newfloat}
\usepackage{listings}
\DeclareCaptionStyle{ruled}{labelfont=normalfont,labelsep=colon,strut=off} 
\floatstyle{ruled}
\newfloat{listing}{tb}{lst}{}
\floatname{listing}{Listing}
\title{A Pioneering Neural Network Method for Efficient and Robust Fluid Simulation}
\author{
    Yu Chen,
    Shuai Zheng\thanks{Corresponding author},
    Nianyi Wang,
    Menglong Jin,
    Yan Chang
}
\affiliations{
    School of Software Engineering, Xi'an Jiaotong University, Xi'an, 710049, China\\
    xjtu\_chenyu@163.com, shuaizheng@xjtu.edu.cn
}

\usepackage{bibentry}

\begin{document}

\maketitle

\begin{abstract}
Fluid simulation is an important research topic in computer graphics (CG) and animation in video games. Traditional methods based on Navier-Stokes equations are computationally expensive. In this paper, we treat fluid motion as point cloud transformation and propose the first neural network method specifically designed for efficient and robust fluid simulation in complex environments. This model is also the deep learning model that is the first to be capable of stably modeling fluid particle dynamics in such complex scenarios. Our triangle feature fusion design achieves an optimal balance among fluid dynamics modeling, momentum conservation constraints, and global stability control. We conducted comprehensive experiments on datasets. Compared to existing neural network-based fluid simulation algorithms, we significantly enhanced accuracy while maintaining high computational speed. Compared to traditional SPH methods, our speed improved approximately 10 times. Furthermore, compared to traditional fluid simulation software such as Flow3D, our computation speed increased by more than 300 times. 
\end{abstract}

%

\section{Introduction}
Fluid simulation intersects multiple disciplines, including computer graphics, 3D computer vision, fluid dynamics, etc. There are two typical representations of fluid in the area of Computational Fluid Dynamics: grid-based methods and particle-based (also known as SPH). Traditional fluid simulations typically use grid-based fluid simulation software such as Ansys Fluent and Flow3D. The grid-based approach segments the fluid region into a grid of cells or voxels. This approach can maintain high numerical stability and has been widely applied in engineering. However, it requires significant computation time and storage, often taking weeks or even months to complete, and it struggles with handling complex boundaries. In recent years, fluid simulation methods based on SPH \cite{solenthaler2009predictive, bender2015divergence, ye2019smoothed, li2018multidisciplinary, koschier2020smoothed} have gained traction in fluid simulation \cite{zheng2021topology, calderon2019sph}, such as DualSPHysics \cite{alshaer2017smoothed, pourabdian2017multiphase}, due to their relatively higher efficiency and better handling of free surface flows. However, despite these advantages over grid-based methods, the inherently high computational cost of solving Navier-Stokes equations remains, which is highly inconvenient for practical applications.

Neural network-based approaches for learning physics have brought new vitality to this field \cite{ling2016reynolds, tompson2017accelerating, morton2018deep, SAHA2021359}. By adopting the particle-based perspective of SPH and integrating methods from 3D computer vision, we treat fluids as point clouds with velocity vectors. This allows us to model complex fluid dynamics using neural networks, thereby avoiding the computationally intensive Navier-Stokes equations and significantly improving computational efficiency.

However, existing neural network methods \cite{ummenhofer2019lagrangian, shao2022transformer, prantl2022guaranteed, li2018learning} are not sufficiently robust, which is suitable only for simple scenarios, such as the freefall of fluid blocks or dam-break flows. In contrast, the internal structure of a tank with ribs and holes is much more complex. Current neural network methods lack sufficient learning capacity and the necessary constraints to guide the modeling of fluid dynamics in such intricate scenarios, making them prone to breakdown under such challenging conditions.

In general, the main contributions of this paper include:
\begin{itemize}
\item[$\bullet$] 
We propose a neural network that is the first to be robust enough to simulate fluid dynamics in complex tanks, achieving a speed improvement of over 300 times compared to traditional fluid simulation software such as Flow3D, and nearly 10 times compared to traditional SPH methods.
\item[$\bullet$] 
We introduce three key capabilities for fluid simulation neural networks: fluid dynamics modeling, physical law constraints, and global stability control. To achieve an optimal balance among these factors, we design the Triangle Feature Fusion, significantly enhancing the accuracy of neural network-based fluid simulation methods.
\item[$\bullet$] 
We have constructed the first complex fluid sloshing surface dataset, encompassing 320,000 frames across four typical tank types and covering various possible flight maneuvers. Compared to existing datasets, this dataset features more complex fluid motions and scenarios, including multi-directional rotations. 
\item[$\bullet$] 
We conducted comprehensive experiments on the fluid dataset and aircraft takeoff generalization scenarios, demonstrating that our network surpasses all previous methods in multiple evaluation metrics.
\end{itemize}

\section{Related Work}
\subsection{Fluid Dynamics Modeling}
When applying point cloud perspectives to SPH fluid particles, an intuitive initial approach might be to model the fluid particles and their interactions using the nodes and edges of a graph neural network \cite{shao2022transformer, li2018learning, battaglia2016interaction, sanchez2020learning, mrowca2018flexible}. However, the complexity of processing large graph structures with numerous nodes and edges makes the convolution operations for each node in GNN highly time-consuming. Some graph-based fluid simulation methods are even slower than the traditional SPH algorithms. Given that the accuracy of current neural network-based methods has not yet surpassed traditional algorithms, it is essential for the computational speed of neural network algorithms to be significantly faster than that of traditional methods to maintain a competitive edge. Graph-based methods rely on particle discretization, but fluid mechanics in the real world are described by continuous partial differential equations rather than discrete graph structures. From this perspective, continuous convolution \cite{Wang_2018_CVPR, ummenhofer2019lagrangian, thomas2019kpconv, winchenbach2024symmetric} are more appropriate than graph convolutions for fluid dynamics modeling. Among these, the CConv \cite{ummenhofer2019lagrangian} has been validated by \cite{prantl2022guaranteed, chen2024dualfluidnet} as a stable foundation for fluid dynamics modeling, enhancing the neural network's ability to learn fluid dynamics.

\subsection{Physical Law Constraints}
Although the continuous convolution kernel can effectively model fluid dynamics, it lacks the constraints of physical laws and additional knowledge. This limitation restricts its accuracy ceiling \cite{cai2021physics}. \cite{prantl2022guaranteed} innovatively applied an anti-symmetric design to CConv, introducing the Anti-Symmetric Convolutional Kernel (ASCC) and demonstrating that this design effectively maintains the law of momentum conservation. However, the hard constraints introduced by the anti-symmetric design reduce ASCC's learning ability for fluid dynamics. To balance this trade-off, the authors introduced a multi-scale feature module, which is relatively time-consuming. In this way, the full potential of CConv and ASCC has not yet been fully explored. In this paper, we propose an innovative feature fusion method that achieves an optimal balance in this trade-off.

\subsection{Global Stability Control}
Besides the local interactions between fluid particles, we also note that learning global fluid features can enhance the neural network's performance in global stability control. Previous works can effectively extract global features from point clouds \cite{qi2017pointnet, qi2017pointnet++, chen2023rotation,wang2019dynamic, wu2024pose, liu2022copy}. In this task, we require a simple and efficient global feature extraction layer, which enables the neural network to broadly control the overall fluid motion within a reasonable range (e.g., fluid particles in a closed box should not move outside the box). The extracted global feature does not need to be highly detailed, as fluid dynamics primarily focuses on the interactions between local particles. The global feature serves as additional information to make the framework more robust.

\begin{figure}
  \centering
    \includegraphics[width=0.9\linewidth]{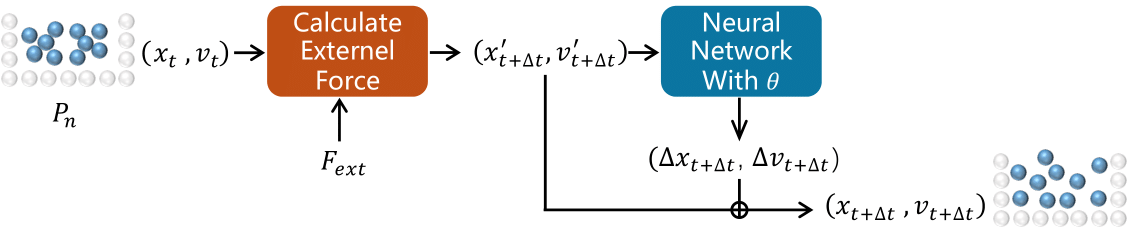}
    \caption{Position-based fluids scheme. It first computes external forces to obtain intermediate fuel particle states. Then, our neural network with trainable parameters \(\theta\) predicts position and velocity changes induced by internal forces.}
    \label{fig:pbf}
\end{figure}

\begin{figure*}
  \centering
    \includegraphics[width=0.91\linewidth]{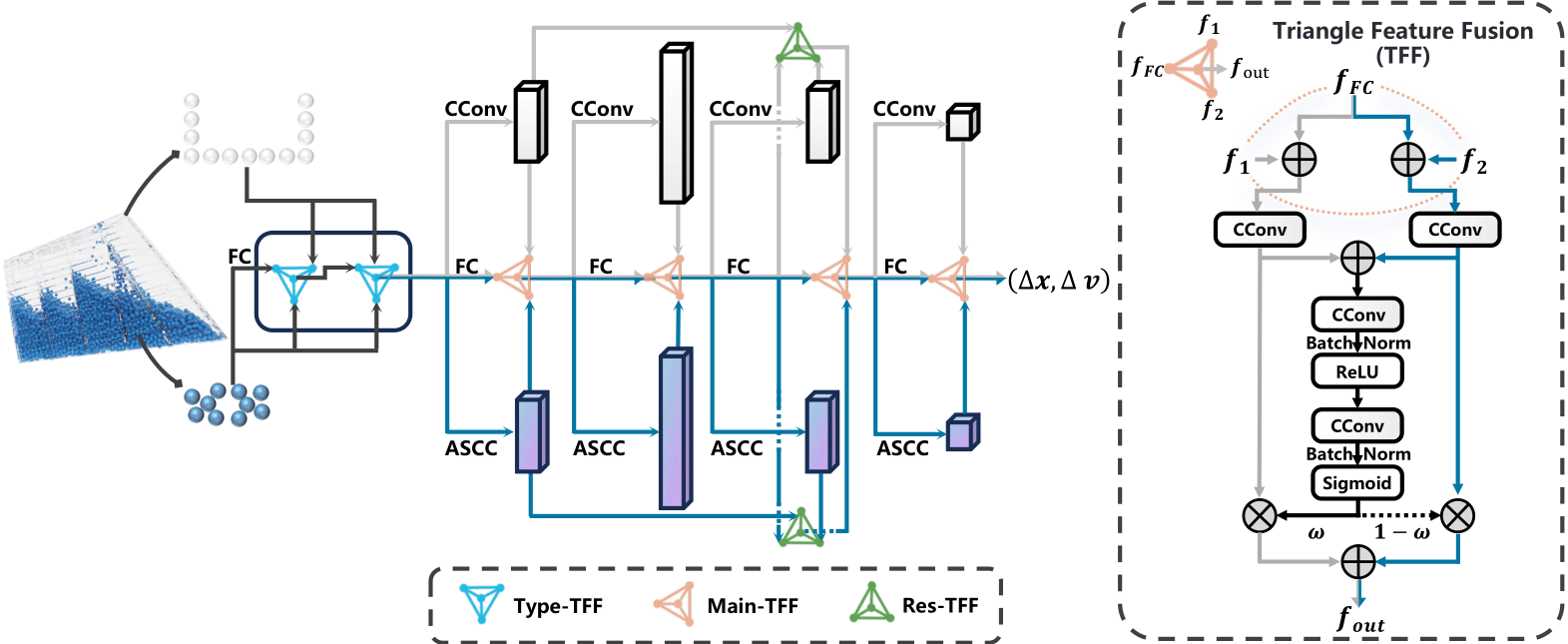}
    \caption{The architecture of our network and Triangle Feature Fusion (TFF). The three types of TFF modules share the same architecture but serve three distinct roles in different positions within the network. The Type-TFF handles type-aware input for fuel and tank particles. The Main-TFF integrates three pathways to balance fluid dynamics modeling, physical constraints, and global stability. The Res-TFF adds a residual connection between the second and fourth layers. }
    \label{fig:network_architecture}
\end{figure*}

\section{Method}
\label{Method}

Our approach employs the position-based fluids (PBF) scheme\cite{macklin2013position, zhang2015position}, as illustrated in Figure \ref{fig:pbf}. Given a set of fuel particles \(P_n\), it takes their positions \(x_t\) and velocities \(v_t\) at time step \(t\) as input, and outputs the final results \(x_{t+1}\) and \(v_{t+1}\) at time step \(t+1\). The main goal of this task is to closely model the physical dynamics, enabling accurate prediction of the particles' positions and velocities at the subsequent time step.

\subsection{Continuous Convolution Kernel}
Inspired by \cite{ummenhofer2019lagrangian, prantl2022guaranteed, chen2024dualfluidnet}, our network incorporates two types of continuous convolution kernels, \textit{CConv} and \textit{ASCC}. For a point cloud \(P_n\) consisting of \(n\) points indexed by \(i\), each corresponding to values \(f_i\) at positions \(x_i\), the \textit{CConv} at position \(x \in P_n\) is described as follows:
\begin{equation}\label{eq2}
    \begin{aligned}
CConv_{g}&=\left ( f*g \right )\left ( x \right ) \\   &=\sum_{i\in \mathcal{N}\left (x,R  \right )}^{}a\left (x_{i}, x  \right )f_{i}g\left ( \Lambda \left ( x_{i} -x \right )  \right ).
    \end{aligned}
\end{equation}
Here, \(\mathcal{N}\left (x, R \right )\) denotes the set of points located within a radius \(R\) centered at position x. The filter function \(g\) employs a mapping function \(\Lambda (r)\) to transform a unit ball into a unit cube. The window function \(a\) scales with the distance between \(x_i\) and \(x\), and becomes zero beyond distance \(R\).

\textit{ASCC}, a specialized variant of \textit{CConv}, uses an antisymmetric kernel to enforce strict constraints, enabling the capture of fluid dynamics while conserving momentum. This is accomplished by halving the learnable kernel parameters along an axis, mirroring the second half about the kernel's center, and then negating the mirrored values. The ASCC formulation is as follows:
\begin{equation}\label{eq4}
    \begin{aligned}
ASCC_{g_{s}}&=\left ( f*g_s \right )\left ( x \right ) \\ &=\sum_{i\in \mathcal{N}\left (x,R  \right )}^{}a\left (x_{i}, x  \right )(f+f_{i})g_s\left ( \Lambda \left ( x_{i} -x \right )  \right ).
    \end{aligned}
\end{equation}
Where \(g_s\) is the anti-symmetric continuous convolution kernel and \(f\) is the corresponding value at position \(x\). For two points \(x\) and \(y\) in \(P_n\), the interparticle force is given by:
\begin{equation}\label{eq:Fxy=Fyx}
    F_{xy}=(f_x+f_y)(-g_s(x-y))=-F_{yx}.
\end{equation}
This ensures that:
\begin{equation}
    \int_{x\in P_n}^{} \int_{y\in P_n}^{} F_{xy}=0.
\end{equation}

The antisymmetric kernel design enforces momentum conservation in \textit{ASCC}, but this constraint limits its learning capacity, complicating the problem. Conversely, \textit{CConv} offers better learning and modeling capabilities but lacks adherence to physical laws. To achieve a balance among these two aspects and global stability control, we propose the Triangle Feature Fusion in the next section.

\subsection{Pipeline with Triangle Feature Fusion (TFF)}
The architecture of our network is shown in Figure \ref{fig:network_architecture}. We propose three types of TFF modules with a shared architecture, as illustrated in the right part of Figure \ref{fig:network_architecture}. However, they serve different roles in different positions within the network. Initially, the fuel particles and tank particles, along with the global fluid features obtained through a fully connected (FC) layer, are input into the Type-TFF. This module is designed to handle the fluid-solid coupling for different types of inputs. 

After passing through two Type-TFF modules, the output features are fed into three separate paths: CConv-based, ASCC-based, and FC. The network architecture consists of five layers. In each layer, features from the three paths undergo fusion through the Main-TFF before being passed to the next layer. This step is crucial for achieving the optimal balance of our three key capabilities. Both CConv and ASCC are based on continuous convolution, providing strong learning and modeling abilities for fluid dynamics. ASCC also introduces a hard constraint for momentum conservation, guiding the network to adhere to physical laws. Additionally, the FC layer extracts global features of the fluid. Incorporating this simple fully connected layer enables the neural network to broadly control the overall fluid motion within a reasonable range (e.g., fluid particles in a closed box should not move outside the box).

Notably, we incorporate a residual connection between the second and fourth layers through the Res-TFF, which benefits the training optimization process.

The architecture of the Triangle Feature Fusion Module is depicted on the right side of Figure \ref{fig:network_architecture}. The features \(f_{FC}\) output from the FC layer are first concatenated with \(f_1\) and \(f_2\) separately. After passing through a \textit{CConv} layer, the two resulting features are concatenated again and pass through several layers of continuous convolutions and ReLU activation. Then a sigmoid function is applied to obtain a fusion weight \(\omega\) between 0 and 1. This weight is used to fuse the features of two pathways, resulting in \(f_{out}\). The mathematical formula is described as follows:
\begin{equation}
    {f_1}' = (f_{1})\oplus(f_{FC}).
 \end{equation}
\begin{equation}
    {f_2}' = (f_{2})\oplus(f_{FC}).
\end{equation}
\begin{equation}\label{eq:w}
    \omega=\lambda (\phi ({f_1}')\oplus \phi ({f_2}')).
\end{equation}
\begin{equation}\label{eq:fusion}
    f_{out}=\omega \times {f_1}'+(1-\omega) \times {f_2}'.
\end{equation}
In Equation \ref{eq:w}, \(\lambda \) comprises two \textit{CConv} layers, \textit{RELU} activation, and the \textit{Sigmoid} function. \(\phi \) represents the \textit{CConv} function(Equation \ref{eq2}). In Type-TFF, \(f_1\) and \(f_2\) represent the fuel particles and the tank particles. In Main-TFF, they represent the features output by the \textit{CConv} layer and the \textit{ASCC} layer. In Res-TFF, they represent the features output by the second and fourth layers.

\subsection{Training Strategy}
\label{training_strategy}
Our loss function is defined as the mean absolute error (MAE) of the position values between the predicted and ground truth, weighted by the neighboring points count:
\begin{equation}\label{eq:loss}
L(t)=\sum_{i}^{ }e^{-\frac{c_{i} }{c_{avg}}}\left \| x_{i}^{t+1} - \hat{x}_{i}^{t+1}  \right \|_{2}^{\gamma}.
\end{equation}
Here, \(x_{i}^{t+1}\) and \(\hat{x}_{i}^{t+1}\) are the predicted and ground truth positions at time step \(t+1\). \(c_i\) is the neighbor count for particle \(i\). \(c_{avg}\) denotes the average neighbor count, set to 40.

Unlike previous methods \cite{chen2024dualfluidnet, ummenhofer2019lagrangian} that compute loss and gradients over the next two frames, our network runs for \(N=W+T\) steps, with each step's input based on the result of the previous prediction. Here, \(W\) represents the preprocessing frames and \(T\) represents the frames over which the loss and gradients are actually computed. During training, the value of \(W\) is adjusted according to training progress and the difficulty of the examples, but loss and gradients are always computed over the \(T\) steps, as formulated in Equation \ref{eq:training_strategy}. This strategy helps the network self-correct over longer sequences, improving long-term prediction stability without increasing memory usage.
\begin{equation}\label{eq:training_strategy}
L_{sum} = \frac{1}{T} {\textstyle \sum_{t=0}^{T}}L(t)  
\end{equation}

\section{Fuel Surface Sloshing Dataset}
\label{Fuel Surface Dataset}

\subsection{Limitation of Existing Datasets}
Existing SPH simulation datasets for fluid simulation \cite{ummenhofer2019lagrangian,chen2024dualfluidnet, shao2022transformer} have the following limitations: 
\begin{itemize}
\item[1)] These datasets include only simple fluid motions, such as free fall or dam break in simple cubic containers, which are inadequate for capturing the fluid dynamics in complex scenarios.
\item[2)] These datasets only focus on scenarios where the system is horizontally placed and subjected to constant vertically downward gravity. However, in real-life scenarios, rotations are common, and forces may come from various directions, especially during the flight of aircraft fuel tanks.
\item[3)] They only focus on water-based simulations. In addition, the fluid motion in fuel tanks often involves surface sloshing rather than fluid blocks falling. This makes existing datasets inadequate for training neural networks to perform well in fuel tank environments.
\end{itemize}

Such simple fluid datasets are inadequate for capturing the complex fuel dynamics in tanks with intricate rib structures. To facilitate a more comprehensive study of aircraft fuel sloshing, it is necessary to construct a dedicated fuel surface dataset. Instead of the commonly used freefall or dam-break scenarios of fluid blocks, we adopt a surface scheme that better represents the actual fuel sloshing inside fuel tanks. Additionally, we employ an iterative strategy to generate data, enriching the initial surface shapes and sloshing scenarios.

\begin{figure}
    \centering
    \includegraphics[width=0.88\linewidth]{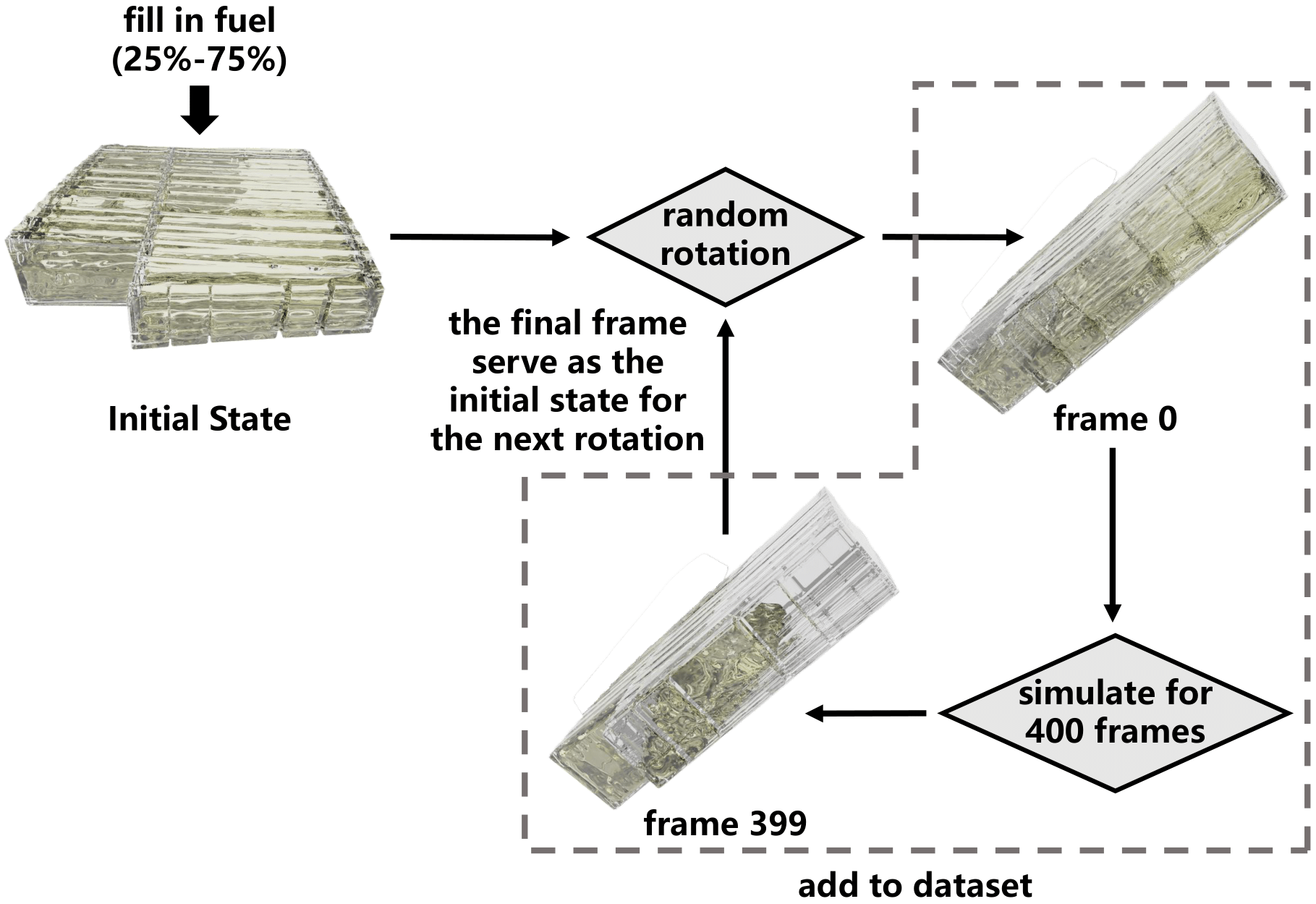}
    \caption{Construction strategy of Fueltank dataset. We use an iterative generation strategy, producing 400 frames per iteration. The final frame of the current iteration serve as the initial state for the next iteration.}
    \label{fig:construction_strategy}
\end{figure}

\renewcommand{\dblfloatpagefraction}{.7}
\begin{figure*}
  \centering
    \includegraphics[width=0.88\linewidth]{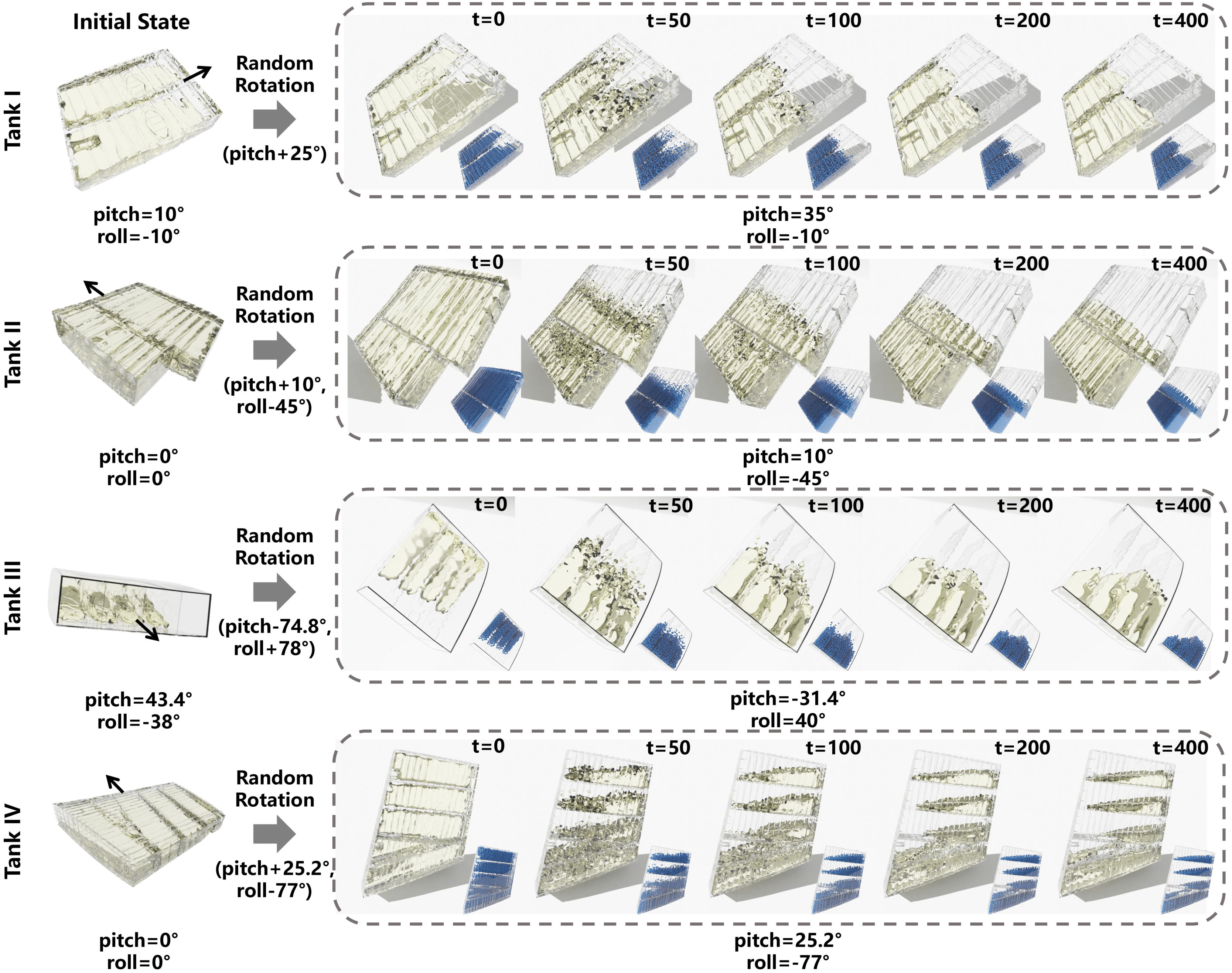}
    \caption{Examples of four tank types from the Fueltank dataset. The fuel tank undergoes random pitch and roll rotations at frame 0 and flows over the next 400 frames. The thumbnails depict the fuel surface from the SPH particle perspective.}
    \label{fig:dataset_alltanks}
\end{figure*}

\subsection{Construction Strategy}
We construct the Fueltank dataset based on four common types of aircraft fuel tanks: Tank Types I, II, III and IV. We generate the SPH fluid dataset using the traditional DFSPH solver \cite{bender2015divergence}, known for its high-fidelity simulations. The particle radius is set to 0.025 \(m\) and the fuel density to 782.885 \(kg/m^3\). 

Our construction strategy is illustrated in Figure \ref{fig:construction_strategy}. In each scenario, we first randomly fill a specific fuel quantity, ranging from 25\% to 75\% of the tank's capacity, in a stationary horizontal tank. Then we apply random rotations to the pitch and roll angles, from -90 to 90 degrees. Notably, these rotations are performed in a very short time, assuming the fuel remains stationary during the rotation. Once the rotation is complete, the fuel begins to flow for 400 frames. We add these frames to the dataset, and the final frame of the current rotation serve as the initial state for the next rotation. We generated 200 iterations for each fuel tank type. This iterative strategy makes our dataset encompass a diverse range of flight maneuvers, which significantly enhances the robustness of neural networks in fuel simulation. Examples from the Fueltank dataset are shown in Figure \ref{fig:dataset_alltanks}.

\section{Experiments}
\label{Experiment}
\subsection{Experimental Setup}
We employ the PyTorch framework and the Adam optimizer for training, using a batch size of 2. The initial learning rate is set to 0.002 and is halved at steps 15000, 25000, ..., 55000. The networks are trained for a total of 60,000 iterations on an NVIDIA A800. Specifically, the particle radius is set to \(h=0.025m\), and spherical filters with a spatial resolution of [4,4,4] are utilized, featuring a radius of \(R=4.5h\).

\subsection{Evaluation Metrics}
Initially, we use the Chamfer Distance (CD) and Earth Mover Distance (EMD) as our evaluation metrics. We calculated the CD and EMD for the next two frames to evaluate the accuracy of the network's short-term prediction.

Additionally, to assess the long-term stability of the fluid simulation, we computed the average distance from the ground-truth particles to the closest predicted particle over the entire sequence of \(n\) frames:
\begin{equation}\label{eq11}
    \begin{aligned}
d^{n} =\frac{1}{N} \sum_{i=1}^{N} \min_{x^n\in X^n}\left \| \hat{x}_i^n -x^n \right \|  _2.
    \end{aligned}
\end{equation}
Where \(X^n\) is the set of predicted particle positions for the frame \(n\), \(\hat{x}_i^n\) is the ground-truth position of particle \(i\), and \(N\) is the total number of particles.

Furthermore, to evaluate the network's adherence to physical laws, we also introduced the maximum density error, which reflects the fluid's incompressibility and stability:
\begin{equation}
     e =\left | 1-\frac{\max_{i} \rho (x_i)}{ \max_{i} \rho ( \hat{x}_i)}  \right | 
\end{equation}

We conducted experiments on our proposed Fueltank dataset, using the traditional SPH method, DFSPH \cite{bender2015divergence}, as ground truth, as it closely approximates real-world fluid dynamics. We compared our method against the traditional methods and existing neural networks.

\renewcommand{\dblfloatpagefraction}{.3}
\begin{table*}
  \centering
    \begin{tabular}{lcccccccc}
      \toprule[0.6pt]
      \multirow{2}{*}{Method} & \multicolumn{2}{c}{CD (mm)} & \multicolumn{2}{c}{EMD (mm)} & \multirow{2}{*}{\shortstack{n-frame Sequence Error\\ \(d^n\) (mm)}} & \multirow{2}{*}{\shortstack{Max Density Error\\  (\(g/cm^{3}\))}} & \multirow{2}{*}{Time (s)} \\ \cline{2-3} \cline{4-5}
                              & t+1 & t+2 & t+1 & t+2 & & & \\ \midrule \midrule
      Grid-based                 & - & - & - & - & - & - &  \textgreater 50\\
      SPH-based                 & - & - & - & - & - & - &  \textgreater 1\\
      CConv                & 1.518 & 3.786 & 0.598 & 0.978 & 160.841 & 0.173 & \textbf{0.021}\\
      DMCF               & 1.472 & 3.488 & 0.185 & 0.323 & 128.902 & 0.025 & 0.602\\
      TIE               & 1.528 & 3.734 & 0.196 & 0.372 & 140.813 & 0.089 & 1.213\\
      DualFluidNet        & 1.232 & 3.169 & 0.157 & 0.328 & 35.318 & 0.013 & 0.199\\
      Ours                   & \textbf{1.149} & \textbf{2.759} & \textbf{0.147} & \textbf{0.259} & \textbf{25.310} & \textbf{0.008} & \underline{0.158} \\ \bottomrule[0.6pt]
    \end{tabular}

  \caption{Quantitative experiments. Grid-based and SPH-based methods are traditional fluid simulation methods. Their simulation results can be regarded as ground truth but require long computation times. Existing neural networks cannot stably simulate fuel sloshing in complex fuel tanks, leading to substantial numerical errors, particularly in long-term predictions. Only our network achieves low errors while significantly reducing computation time.}
  \label{tab:Quantitative_experiments}
\end{table*}

\begin{figure}[ht]
  \centering
    \includegraphics[width=0.885\linewidth]{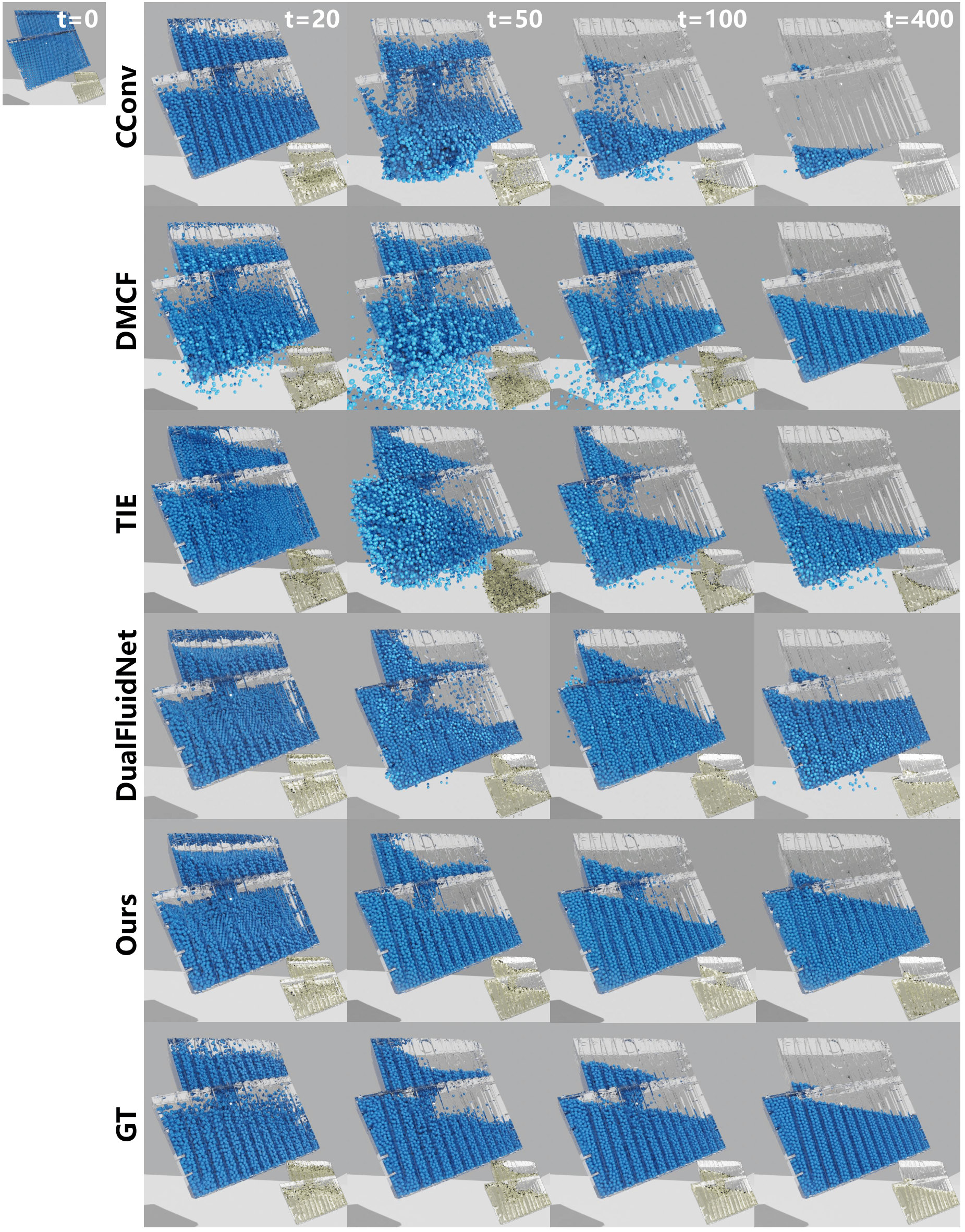}
    \caption{Qualitative experiments on Tank II show that existing neural network methods fail in complex scenarios, whereas our method provides stable fluid simulations comparable to traditional methods.}
    \label{fig:Qualitative_experiments}
\end{figure}

\subsection{Experiments on Fueltank dataset}
The results of the comparative experiments are shown in Table \ref{tab:Quantitative_experiments} and Figure \ref{fig:Qualitative_experiments}. Existing methods struggle in complex scenario. Specifically, the qualitative results of the continuous convolution methods, CConv \cite{ummenhofer2019lagrangian} and DMCF \cite{prantl2022guaranteed}, show particles falling vertically continuously. This occurs because they overfit to a specific gravity direction and cannot handle varying rotational angles, causing the networks' predictions for internal forces ultimately converge to zero in all directions. According to the PBF scheme, as discussed in Figure \ref{fig:pbf}, when the internal forces are zero, the fluid moves downward due to gravity.

We also conducted experiments on the current state-of-the-art methods: TIE \cite{shao2022transformer}, based on graph convolution, and DualFluidNet \cite{chen2024dualfluidnet}, based on continuous convolution. TIE's inadequate fluid dynamics modeling causes particles to disperse erratically during simulation. Additionally, the large number of nodes and edges processed by the graph convolution and the incorporation of transformers result in significant computation time. For DualFluidNet, although it is the most stable among the existing methods, it still experiences particles escaping the fuel tank. Furthermore, since DualFluidNet computes loss and gradients only over the next two frames, its long-term stability is inferior to our network's W+T training strategy, as discussed in \ref{training_strategy}.

\begin{figure}[t]
  \centering
  \subfigure[Tank type I]{
    \includegraphics[width=0.92\linewidth]{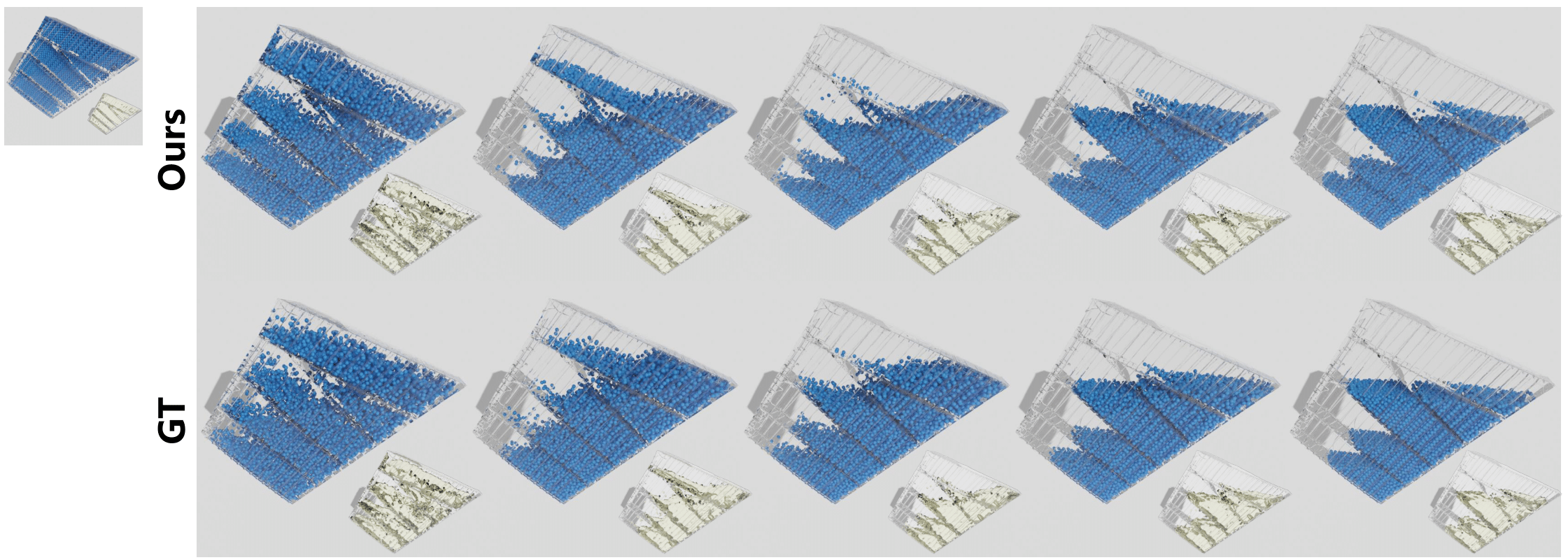}}
  \subfigure[Tank type III]{
    \includegraphics[width=0.92\linewidth]{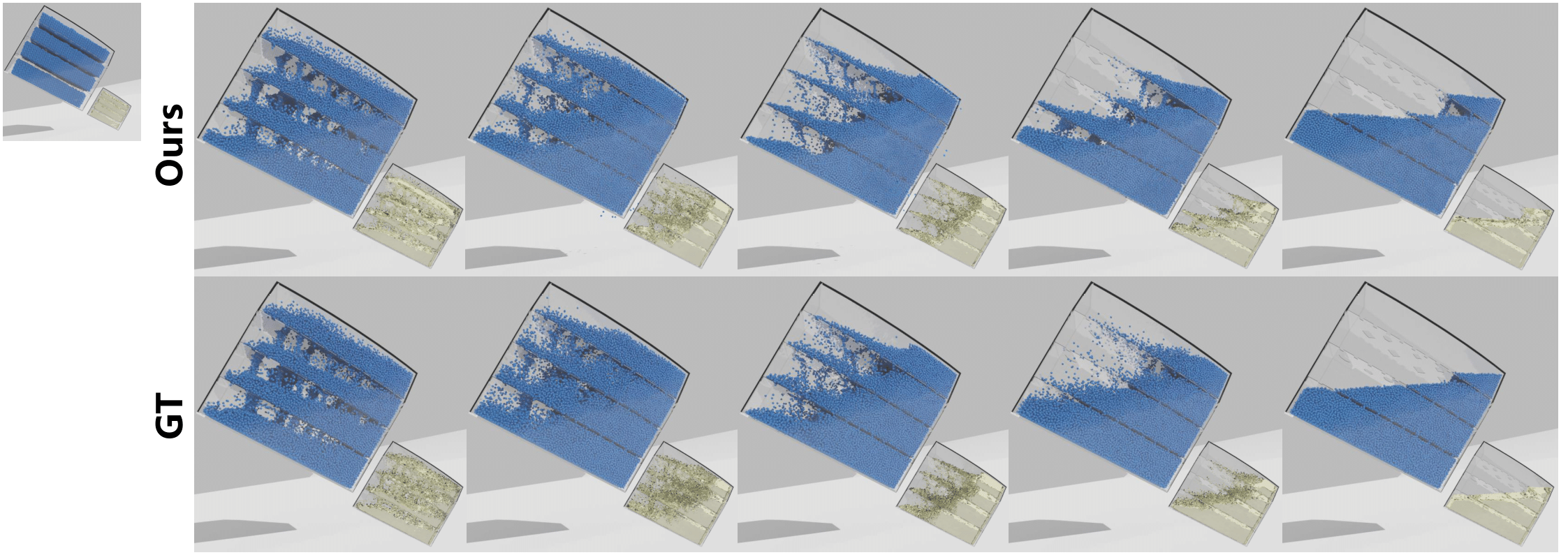}}
  \subfigure[Tank type IV]{
    \includegraphics[width=0.92\linewidth]{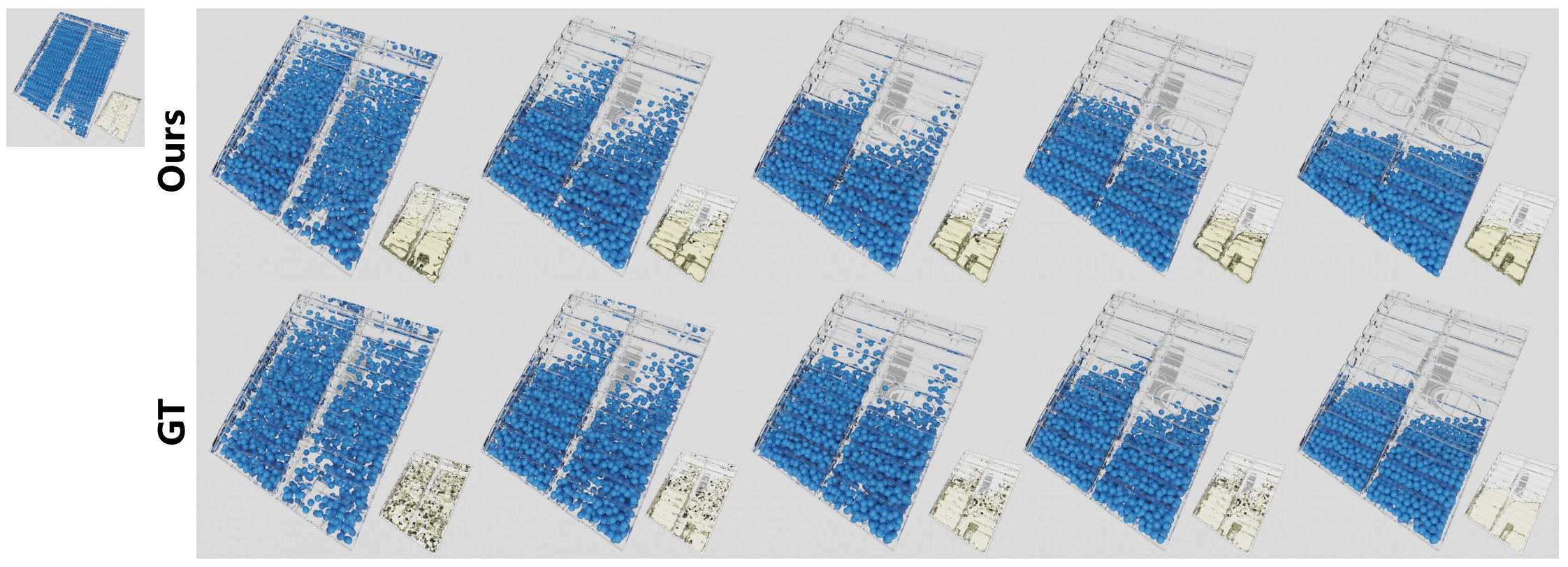}}
  \caption{Qualitative experiments on other three tanks.}
  \label{fig:other tanks}
\end{figure} 

Our method is the first neural network robust enough to simulate fuel sloshing in complex fuel tanks, achieving accuracy comparable to traditional methods while significantly increasing speed. Like the stability of a triangle, our proposed triangle feature fusion achieves an optimal balance among the three key capabilities, providing exceptionally stable fluid simulation capabilities. Figure \ref{fig:other tanks} shows the performance of our network in the other three fuel tanks.

\begin{figure*}[ht]
  \centering
    \includegraphics[width=\linewidth]{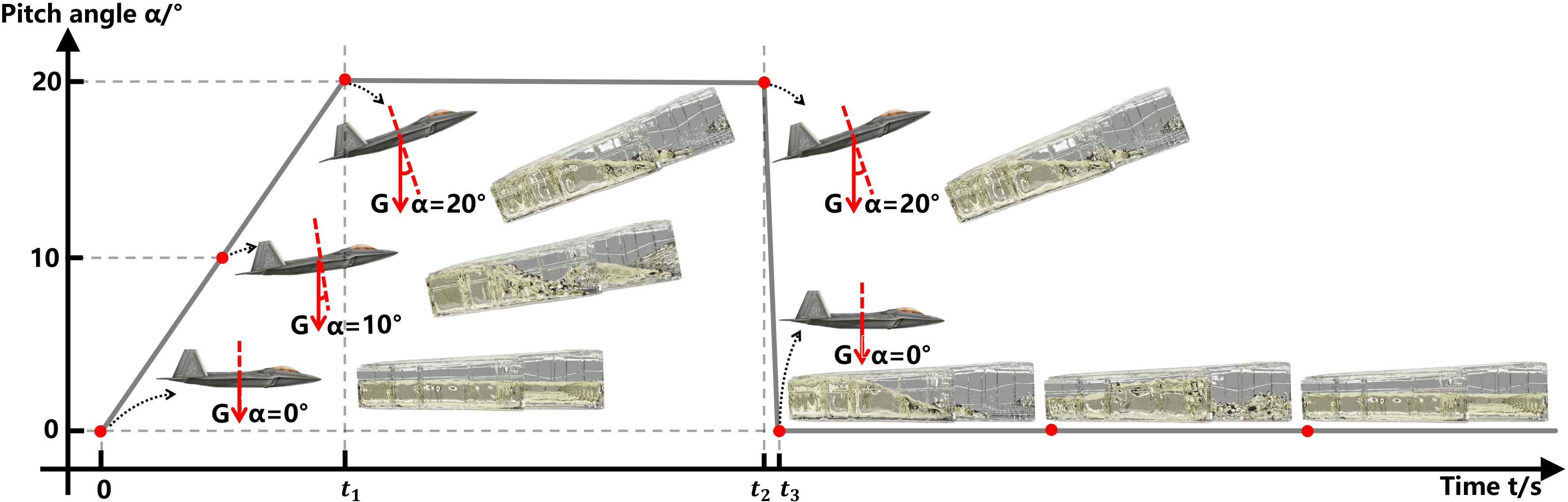}
    \caption{We simulated the aircraft's takeoff process, where the pitch angle gradually increases from 0 to 20 degrees at a pitch rate of 2 degrees per second from \(t_0\) to \(t_1\). It maintains 20 degrees from \(t_1\) to \(t_2\) , and then rapidly returns to 0 degrees from \(t_2\) to \(t_3\). From \(t_2\) to \(t_3\), which is a very short period, the fuel surface can be assumed as unchanged.}
    \label{fig:fuel_sloshing_simulation}
\end{figure*}

\begin{table}[t]
  \centering
  \begin{tabular}{lcc}
    \toprule[0.6pt]
    \multirow{2}{*}{Method} & Sequence Error & Max Density Error \\
    & \(d^n\) (mm) & (\(g/cm^{3}\)) \\
    \midrule \midrule
    w/o Main-TFF & 149.521 &  0.098\\
    w/o Type-TFF & 99.237 & 0.032 \\
    w/o Res-TFF & 39.717 &  0.024\\
    w/o CConv & 181.518 & 0.144 \\
    w/o ASCC & 87.922 &  0.054 \\
    w/o FC & 33.317 &  0.013\\
    Ours & \textbf{25.310} & \textbf{0.008} \\
    \bottomrule[0.6pt]
  \end{tabular}
  \caption{Quantitative results of the ablation study, showing the contributions of each TFF module type and the integration of the three features in the Main-TFF module.}
  \label{tab:Ablation_Study}
\end{table}

\renewcommand{\dblfloatpagefraction}{.3}
\begin{figure}
  \centering
    \includegraphics[width=\linewidth]{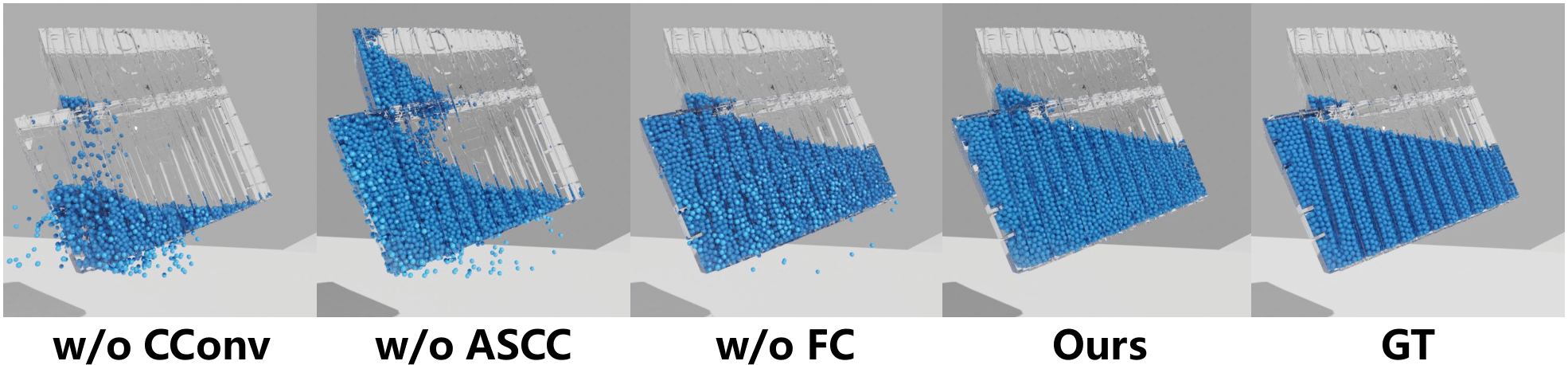}
    \caption{Qualitative results of the ablation study, further validating the necessity and contributions of our triangle feature fusion to network performance.}
    \label{fig:ablation_study}
\end{figure}
\subsection{Ablation Study}

We conducted an ablation study to verify the effectiveness of each module, as shown in Table \ref{tab:Ablation_Study} and Figure \ref{fig:ablation_study}. Specifically, in the w/o CConv scenario, replacing CConv reduced the network's ability to model fluid dynamics, resulting in fluid collapse. In the w/o ASCC scenario, replacing ASCC retained some learning and modeling capability but lacked the guidance of physical laws, leading to deviations from normal physical behavior. Comparing w/o FC and Ours, adding the FC layer can improve global fluid control stability. The Type-TFF enhances the capability to handle type-aware input for fuel and tank particles. The Main-TFF integrates three pathways to balance fluid dynamics modeling, physical constraints, and global stability. The Res-TFF adds a residual connection to benefit the training optimization. Each module has a distinct and indispensable role, collectively forming an efficient and robust framework.

\subsection{Fuel Sloshing Simulation in Flight}
\begin{figure}
  \centering
    \includegraphics[width=\linewidth]{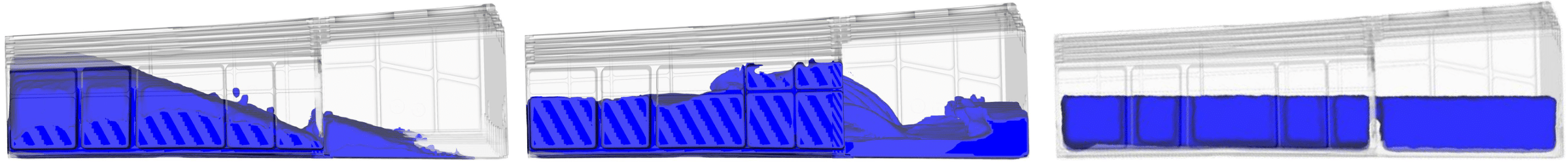}
    \caption{Simulation results after \(t_3\) using traditional fluid simulation software Flow3D, which employs the traditional grid-based method.}
    \label{fig:flow3d}
\end{figure}
To validate the generalization of our method in real aircraft takeoff scenarios, we simulated the aircraft’s takeoff process, where the pitch angle gradually increases from 0 to 20 degrees, stabilizes, and then rapidly returns to 0 degrees, as shown in Figure \ref{fig:fuel_sloshing_simulation}. It shows that our network can generalize to real-world fuel simulation scenarios robustly and stably.  Additionally, we compared the simulation results after \(t_3\) using the famous fuel simulation software Flow3D, as shown in Figure \ref{fig:flow3d}. For the entire simulation, Flow3D takes nearly 10 hours, while our method requires only 2 minutes and achieves comparable accuracy. 

\section{Conclusion}
In this paper, we proposed the first neural network robust enough for fuel sloshing simulation in complex fuel tanks. Our approach optimally balances fluid dynamics modeling, physical law constraints, and global stability control through a triangle feature fusion. We also constructed the first comprehensive fuel surface sloshing dataset. Our network significantly enhances simulation speed while ensuring accuracy and stability in aircraft applications. This network can also generalize to other fluids and scenarios, providing a robust and efficient framework for fluid simulation.

\appendix
\section{Acknowledgments}
The work reported in this paper is supported by the National Natural Science Foundation of China [52375514].

\bibliography{aaai25}

\end{document}